\documentclass[10pt,twocolumn,letterpaper]{article}

\usepackage{iccv}
\usepackage{times}
\usepackage{epsfig}
\usepackage{graphicx}
\usepackage{amsmath}
\usepackage{amssymb}
\usepackage{color}
\usepackage{algorithm}
\usepackage{algorithmic}
\usepackage{multirow}
\usepackage{diagbox}

\usepackage[pagebackref=true,breaklinks=true,letterpaper=true,colorlinks,bookmarks=false]{hyperref}

\iccvfinalcopy 

\def\iccvPaperID{2389} 

\ificcvfinal\fi

\begin{document}

\title{Learnable Expansion-and-Compression Network for \\Few-shot Class-Incremental Learning}

\author{Boyu Yang$^1$ \and Mingbao Lin$^2$ \and Binghao Liu$^1$ \and Mengying Fu$^1$ \and Chang Liu$^1$ \and Rongrong Ji$^{2,3,4}$ \and Qixiang Ye$^{1,4}$\thanks{Corresponding Author.}
\and PriSDL, EECE, University of Chinese Academy of Sciences$^1$ \and MAC, Department of Artificial Intelligence, School of Informatics, Xiamen University$^2$ \and Institute of Artificial Intelligence, Xiamen University$^3$ \and Peng Cheng Laboratory$^4$
\and{\tt\small yangboyu18@mails.ucas.ac.cn}
\and{\tt\small lmbxmu@stu.xmu.edu.cn}
\and{\tt\small \{liubinghao18, fumengying19, liuchang615\}@mails.ucas.ac.cn}
\and{\tt\small rrji@xmu.edu.cn}
\and{\tt\small qxye@ucas.ac.cn}
}

\maketitle
\ificcvfinal\thispagestyle{empty}\fi

\begin{abstract}
Few-shot class-incremental learning (FSCIL), which targets at continuously expanding model's representation capacity under few supervisions, is an important yet challenging problem. 
On the one hand, when fitting new tasks (novel classes), features trained on old tasks (old classes) could significantly drift, causing catastrophic forgetting. 
On the other hand, training the large amount of model parameters with few-shot novel-class examples leads to model over-fitting.
In this paper, we propose a learnable expansion-and-compression network (LEC-Net), with the aim to simultaneously solve catastrophic forgetting and model over-fitting problems in a unified framework.
By tentatively expanding network nodes, LEC-Net enlarges the representation capacity of features, alleviating feature drift of old network from the perspective of model regularization.
By compressing the expanded network nodes, LEC-Net purses minimal increase of model parameters, alleviating over-fitting of the expanded network from a perspective of compact representation.
Experiments on the CUB/CIFAR-100 datasets show that LEC-Net improves the baseline by 5$\sim$7\% while outperforms the state-of-the-art by 5$\sim$6\%. 
LEC-Net also demonstrates the potential to be a general incremental learning approach with dynamic model expansion capability.
Code is anonymously available at \href{https://github.com/Yang-Bob/LECNet}{\color{magenta}github.com/Yang-Bob/LECNet}. 
\end{abstract}

\begin{figure}[t]
    \centering
    \includegraphics[width=1\linewidth]{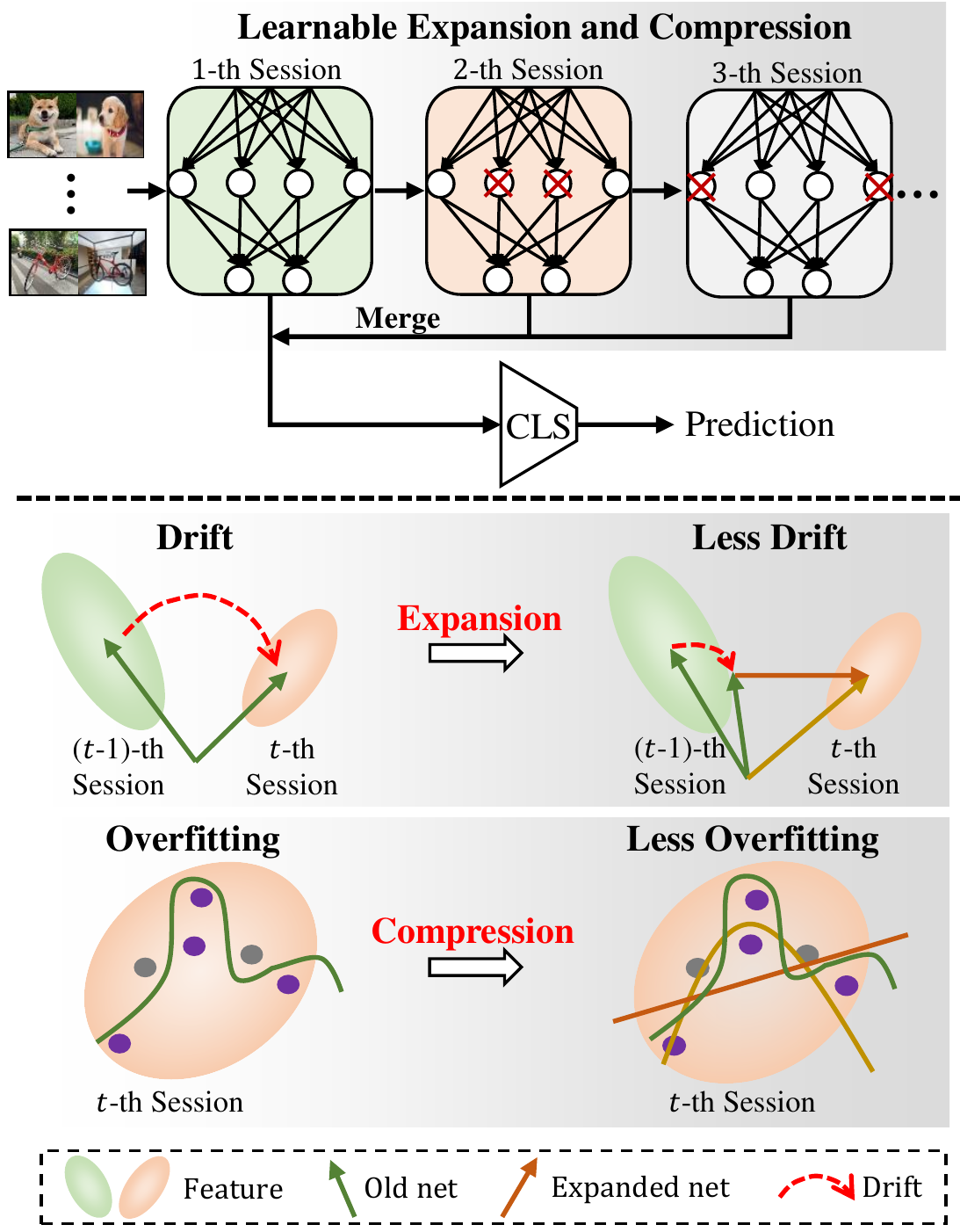}
    \caption{Comparison of conventional methods and our learnable expansion-and-compression network (LEC-Net) for FSCIL. By dynamically expanding the network nodes, LEC-Net enlarges the representation capacity of network, alleviating feature drift by model regularization. By compressing the expanded network nodes, LEC-Net purses adaptive increase of model parameters, alleviating over-fitting from a perspective of compact representation. }
    \label{fig:motivation}
\end{figure}
\section{Introduction}\label{introduction}
%
In the past few years, we witnessed the great progress of visual recognition~\cite{ResNet2016}. This attributes to the availability of large-scale datasets with precise annotations and convolutional neural networks (CNNs) capable of absorbing the annotation information. However, annotating a large amount of objects is laborious and expensive. It is also not consistent with cognitive learning, which not only can build a precise model using few-shot examples but also can generalize the model to novel things in an incremental fashion~\cite{CompositionalRepre-ICCV2019}.  

To improve the generalization capacity of visual recognition models, few-shot class incremental learning (FSCIL)~\cite{TOPIC} is proposed to simulate the computational mechanism of visual cognition. Given base classes with sufficient training data and novel classes of few supervisions, FSCIL trains a representation model from the base classes and continually adapts it to the novel classes. Both base classes and novel classes required to be recognized during inference. 

With above settings, FSCIL faces challenges beyond both few-shot learning and continuous learning. On the one hand, training networks with novel class samples would cause the drift of feature representation, $i.e.$, catastrophic forgetting base/old classes. On the other hand, large-scale models ($e.g.$, deep CNNs) are prone to be overfitting when continuously fintuned with limited examples, Fig.~\ref{fig:motivation}.
The neural gas (NG) method~\cite{TOPIC} pioneered the research in this domain, by preserving the topology of feature manifold. Through feature topology preservation, it regularizes the network training and shows significance to handle the catastrophic forgetting issue. In NG, however, the representation capacity of a fixed number of network nodes experiences difficulty to handle continuously incremental classes.

In this paper, we propose a Learnable Expansion-and-Compression Network (LEC-Net) to solve the challenging FSCIL problem. Given incremental novel classes with unknown distribution, LEC-Net defines a {tentative optimization mechanism}, which first expands network nodes to construct redundant representation capacity, then reduces nodes to obtain compact feature representation for novel classes.
In each training iteration, the node outputs are used to produce an indicator vector by applying an non-linear activation function on the node outputs themselves. When an activated indicator element is equal to 0, the node is removed; otherwise, the node is reserved. In this way, the expanded network nodes are compressed in a self-activated fashion.

LEC-Net aims to simultaneously solve catastrophic forgetting and model over-fitting problems based on the dynamic node expansion module, Fig.~\ref{fig:motivation}. On the one hand, LEC-Net tentatively increases the dimensionality of features and distribute the gradient from old network nodes, alleviating feature drift and catastrophic forgetting by posing strong model regularization to the old network. LEC-Net then adaptively reduce the feature dimensonality so that the gradient of the few-shot incremental samples can focus on learning few features, alleviates over-fitting of the expanded network from the perspective of compact representation. 

To summarize, the contributions of this paper include:

\begin{itemize}
    \item We propose a learnable expansion-and-compression network (LEC-Net), which solves the catastrophic forgetting and model over-fitting problems of FSCIL in a unified framework.  

    \item We provide a self-activation module for dynamic network node expansion and compression, implementing class incremental learning in an adaptive fashion.
    
    \item We improve the state-of-the-art of FSCIL by significant margins, as well as validating the effectiveness of LEC-Net on general incremental learning problems.
\end{itemize}

\section{Related Work}

\textbf{Few-shot Learning.}
Few-shot learning aims to learn a model given sufficient training data from base classes and few supervisions from novel classes. Existing methods can be coarsely categorized into metric learning, meta learning and data augmentation methods. Metric learning methods~\cite{MatchingNet,PrototypicalNet,RelationNet,DeepEMD,PMMs,Harmonic} train two-branch networks to predict whether two images/regions belong to the same category.
Meta learning methods~\cite{MAML,MetaNeural,Meta-Transfer} purses the faster adaptation of model parameters to the new categories with few images. Data augmentation methods~\cite{SaliencyHallucination,AdversarialHallucination,BoundaryAdversarial} generate examples of rich transforms for unseen categories.

Existing studies improved the performance on few-shot novel classes; however, the performance of base classes often significantly degenerated. To solve, Gidaris~\emph{et al}.~\cite{DynamicFewshot} introduced the incremental few-shot learning which forces the model to pay attention to not only novel classes but also the base classes. 
A few-shot classification weight generator based on the attention mechanism~\cite{MatchingNet} was proposed to enhance the performance of base classes when fine-tuning the networks with novel class data. Meta-learning~\cite{AttentionAttractor,XtarNet} and feature alignment methods~\cite{DiscriminantAlignment} were explored to regularize the learning procedure of novel classes.  

\textbf{Incremental Learning}. This line of research can be categorized to task-incremental learning and class-incremental learning (CIL)~\cite{IncrementalSurvey}. 
Task-incremental learning methods can be further categorized to rehearsal ~\cite{iCaRL,RWalk,LargeScaleIL,DeepReplay,ConditionalAdversarial}, regularization ~\cite{LwF,LwM,PathInt,GradCAM,RWalk,LessForget}, and architecture configuration ones~\cite{PackNet,HardAttention,DEN,Piggyback}. Rehearsal methods used a number of exemplars reserved in the previous task or generated some synthetic images/features, and then replayed them in the current task to prevent the forgetting of previous ones. Regularization methods introduced regularization loss functions when training the network. For example, PathInt~\cite{PathInt} and RWalk~\cite{RWalk} considered the weight regularization to prevent feature drift. LwF~\cite{LwF} and LFL~\cite{LessForget} used data regularization to constrain network outputs. Architecture configuration methods designed task-adaptive networks, $e.g.$, hard attention network~\cite{HardAttention}, prunning and pack mechanisms~\cite{PackNet} to choose network parameters during inference. Dynamic explanation network~\cite{DEN} improved network parameters using a three-step strategy including selective retraining, expansion and splitting. Our study not only inherits the advantages of dynamic explanation network, but introduces self-activation module to perform network compression and handle model overfitting.

When task IDs are not accessible during inference, task-incremental learning evolves to class-incremental learning, where only the training data for a number of classes has to be present at the same time and new classes are added progressively~\cite{iCaRL}. The primary challenge for class-incremental learning is catastrophic forgetting, which has been elaborated by various methods including learning without forgetting~\cite{LwF,IncrementalDet2017}, memory schemes~\cite{LwM,Mnemonics2020}, and transfer strategies~\cite{Transfer19}.

\textbf{Few-shot Class Incremental Learning}. It requires to learn new classes with very few labelled samples without forgetting the previously learned ones. Compared with incremental learning, FSCIL faces the serious over-fitting problem brought by few-shot training examples. The neural gas method~\emph{et al}.~\cite{TOPIC} resolved this problem by constructing and preserving the feature topology but remains challenged by the network capacity problem. The dynamic few-shot learning method~\cite{DynamicFewshot} proposed an attention based classification weight generator, which leads to feature representations that generalize better on ``unseen” categories. Nevertheless, this method does not involve network expansion, which limits its potential to a large amount of novel classes.

\section{Methodology}
\subsection{Preliminary}
FSCIL is defined upon base classes $C_{\text{base}}$ of sufficient training data and novel classes $C_{\text{novel}}$ with few supervisions from a stream dataset $\{D^{(t)}, t=0,1,2,...\}$ where $D^{(t)}$ corresponds to the $t$-th incremental class set $C^{(t)}$. For $\forall_{t_1 \neq t_2}$, we have $C^{(t_1)} \cap C^{(t_2)}=\varnothing$, $C^{(0)} = C_{\text{Base}} $, and ${\cup}_t C^{(t)} = C_{\text{Novel}}$ with $t>0$. In the $t$-th session, FSCIL trains the network solely upon the dataset $D^{(t)}$ of class set $C^{(t)}$ and test on all of the seen class $\{C^{(0)}, ..., C^{(t)}\}$, without forgetting the old classes $\{C^{(0)},\cdots,C^{(t-1)}\}$.

A naive solution to handle the FSCIL problem is first to train a network with $C_{\text{base}}$, then fine-tune the network using incremental data $D^{(t)}$ with the novel class set $C^{(t)}$. For the base training session ($0$-$th$ session), the image $\boldsymbol x \in D^{(0)}$ is fed to a convolutional neural network (CNN), and then the feature vector is extracted as
$f(\boldsymbol x; \theta_m) \in \mathbb{R}^{c}$, where $f(\cdot)$ is the convolutional network with its parameters $\theta_m$. Denote $g(\cdot)$ parameterized by $\theta_c^{(0)}$ as the classifier, we have the network prediction $\boldsymbol{\hat{y}}^{(0)} = g\big(f(\boldsymbol x; \theta_m); \theta_c^{(0)}\big)$. Given the ground-truth $\boldsymbol y$ of the image, the network is trained to optimize the following classification loss function:
\begin{align}
   \mathop{\arg\min}_{\theta} \mathcal{L}_{c}(\boldsymbol{\hat{y}}^{(0)}, \boldsymbol y; \theta),
   \label{eq:loss_Base}
\end{align}
where $\theta = \{\theta_m, \theta_c\}$, $\theta_c = \theta_c^{(0)}$, and $\mathcal{L}_{c}=\boldsymbol{y}\log(\boldsymbol{\hat{y}}^{(0)})$ denotes the cross entropy loss function.

\begin{figure}[t]
    \centering
    \includegraphics[width=1\linewidth]{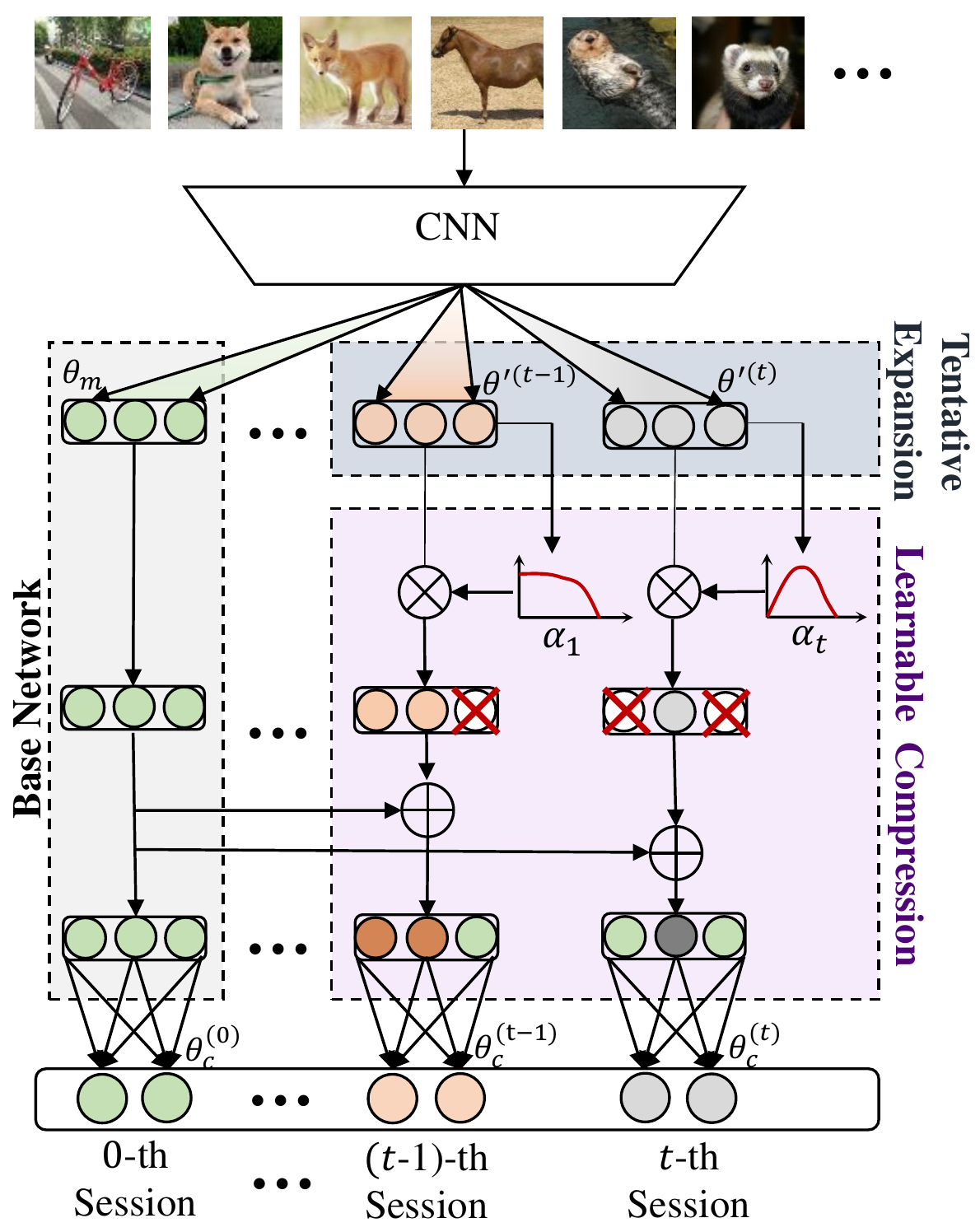}
    \caption{LEC-Net architecture. Given incremental novel classes with unknown distribution, LEC-Net tentatively expands network nodes to construct redundant representation capacity. It then reduces the expanded network nodes to obtain compact feature representation for novel classes.}
    \label{fig:flowchart}
\end{figure}

For the incremental learning sessions (\emph{i.e.}, $t>0$), to classify new classes $C^{(t)}$, more classification parameters $\theta_c^{(t)}$ are added to the network. The prediction of the network becomes $\boldsymbol{\hat{y}}^{(t)} =  g\big(f(\boldsymbol x; \theta_m); \theta_c\big)$, where $\theta_c = \{\theta_c^{(0)},..., \theta_c^{(t)}\}$. It was observed that the shared feature representation $f(\boldsymbol x; \theta_m)$ tends to forget the old classes since the new class data might have very poor sampling within the old class domain~\cite{LwF}. To solve, a distillation loss is adopted to maintain the parameters learnt from the old classes, formulated as $\mathcal{L}_d = \boldsymbol{\hat{y}}^{(t-1)} \log \boldsymbol{\hat{y}}^{(t)}$, where $\boldsymbol{\hat{y}}^{(t-1)}$ is the output of the network trained in the $(t-1)$-th session. With the above definitions, the FSCIL network is trained by
\begin{align}
   \mathop{\arg\min}_{\theta} \mathcal{L}_{c}(\boldsymbol{\hat{y}}^{(t)},\boldsymbol y; \theta)+\lambda_1 \mathcal{L}_d(\boldsymbol{\hat{y}}^{(t)},\boldsymbol{\hat{y}}^{(t-1)}; \theta),
   \label{eq:loss}
\end{align}
where $\theta = \{\theta_m, \theta_c\}$. For base class training, $\theta_c = \{\theta_c^{(0)},...,\theta_c^{(t)}\}$. $\lambda_1$ is a regularization factor.

Despite its capability in learning from large-scale training data, the distillation is challenged by the catastrophic forgetting problem when applied to FSCIL because a fixed number of network nodes experiences the difficulty to remember old classes as well as fitting novel classes. In what follows, we introduce the LEC-Net to solve this problem.

\subsection{Network Expansion-and-Compression}
Although the network performance is observed to be enhanced by expanding network nodes, it is unclear how many nodes should be added in each session. On the one hand, adding few nodes could not increase the representation capacity; on the other hand, adding many nodes leads to significant feature dimensionality increase which aggregates over-fitting problem and degenerates test performance. We propose the tentative strategy, which first adds redundant network nodes for sufficient representation and then adaptively removes the added nodes for compact representation.

\textbf{Tentative Expansion.} In FSCIL, training the network using incremental classes causes the catastrophic forgetting of old classes. To dive into a detailed analysis, when the network is fine-tuned by new classes without the constraint of old classes, network parameters driven by novel gradient quickly drift to a new domain. We propose to tentatively expand the network nodes,  enlarging network representation capacity, Fig.\ \ref{fig:flowchart}.

In specific, the new features generated by the expanded nodes are calculated as $f(\boldsymbol x; \theta_m, \theta^{'(t)})$ where $\theta^{'(t)}$ denotes the expanded network parameters for the novel classes.
The old and expanded features are fused as $f(\boldsymbol x; \theta_m, \theta^{'(t)}) \oplus \big(\gamma f(\boldsymbol x; \theta_m)\big)$, where $\oplus$ denotes the fusing operation such as plus or concatenation and $\gamma$ is a coefficient to balance the two kind of features. Accordingly, the prediction results of the network are rewritten as $\boldsymbol{\hat{y}}^{(t)} =  g(f(\boldsymbol x; \theta_m, \theta^{'(t)}); \theta_c)$ and the expanded network is trained by
\begin{align}
   \mathop{\arg\min}_{\theta, \theta^{'}} \mathcal{L}_{c}(\boldsymbol{\hat{y}}^{(t)},\boldsymbol y; \theta, \theta^{'})+\lambda_1 \mathcal{L}_d(\boldsymbol{\hat{y}}^{(t)},\boldsymbol{\hat{y}}^{(t-1)}; \theta, \theta^{'}),
   \label{eq:loss_exp}
\end{align}
where $\theta^{'} = \theta^{'(t)}$ denotes the expanded network parameters.

Merging the old network with the expanded nodes increases the representation capacity. The rationale behind this is that the added nodes absorb part of the gradient in learning the new classes, all of which are supposed to flow to the old network, which alleviates feature drift and enhances the representation capacity. Besides, the old network trained on abundant training data of base classes, provides a good startup for training the expanded network. Nevertheless, unconstrained expansion of network parameters aggravates overfitting, which is solved by the network compression procedure. 

\begin{figure}[t]
    \centering
    \includegraphics[width=1\linewidth]{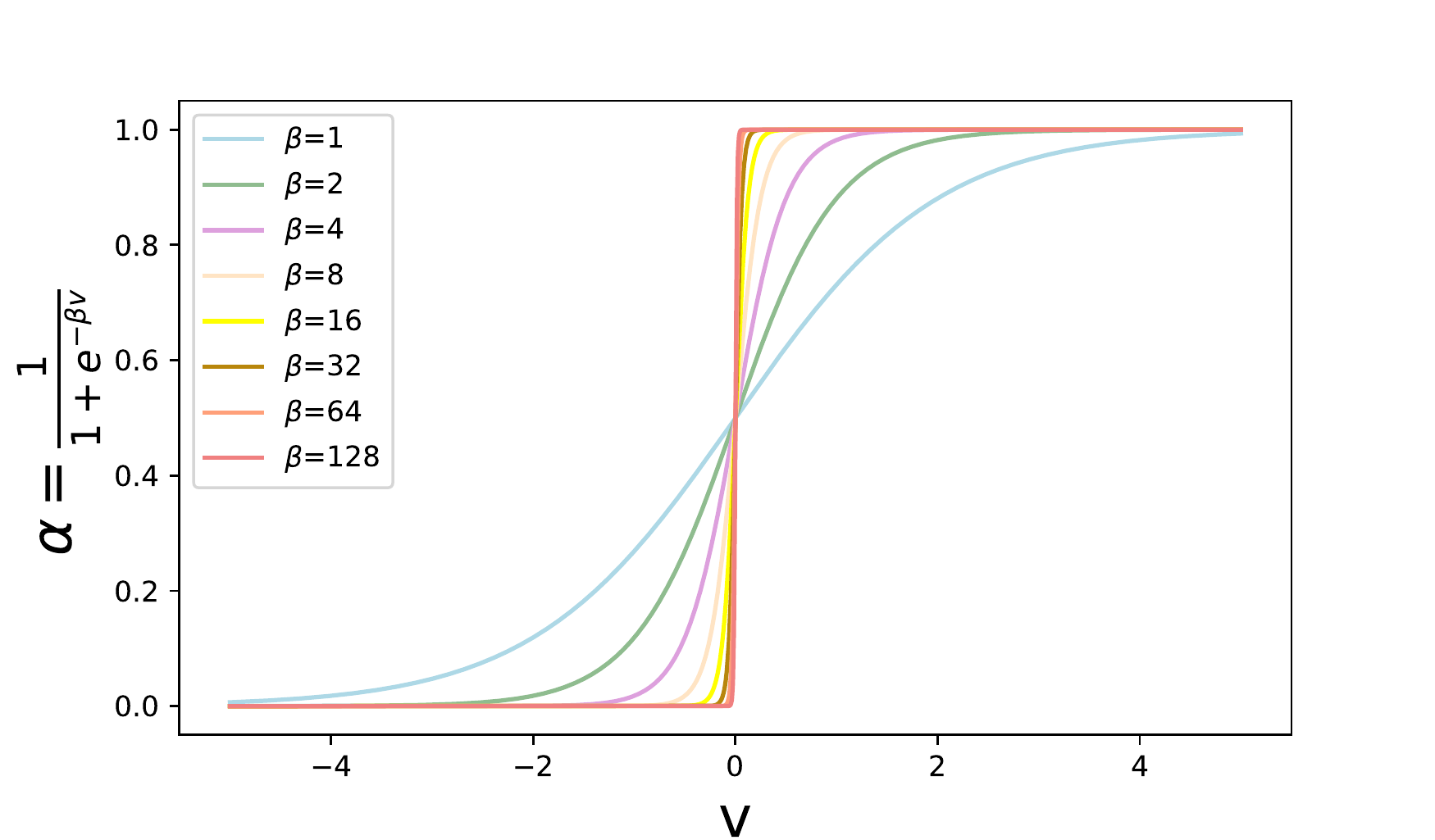}
    \caption{$\alpha-v$ curve upon $\beta$.}
    \label{fig:beta}
\end{figure}

\textbf{Learnable Compression}.
To adaptively remove network nodes, an variable indicating which nodes are informative for the new classes is introduced. With the indicator variable, the features are retained or removed as:
\begin{align}
   \big(\boldsymbol \alpha_t \odot f(\boldsymbol x; \theta_m, \theta^{'(t)})\big) \oplus \big(\gamma f(\boldsymbol x; \theta_m)\big),
   \label{eq:compress}
\end{align}
where $\odot$ denotes the Hadamard product (element-wise multiplication). $\boldsymbol \alpha_t \in \{0,1\}^c$ is an indicator vector satisfying $\|\boldsymbol \alpha_t\|_1 /c < \tau_t$ where $\tau_t$ denotes the node retention rate and $\|\cdot\|_1$ denotes the $\ell_1$-norm. By tuning $\tau_t$, the upper bound of the added node number $\|\boldsymbol \alpha_t\|_1$ is limited, which significantly reduces the network complexity and relieves the over-fitting problem.

During training, we further introduce a learnable parameter $\boldsymbol s_t \in \mathbb{R}^c$ to regularize $\boldsymbol \alpha_t$ as $\boldsymbol \alpha_t = \frac{1}{1+e^{-\boldsymbol s_t}}$. To guarantee that $\boldsymbol \alpha_t$ is sparse, two regularization losses are introduced: (1) $\mathcal{L}_1 = \big\||\boldsymbol s_t|-N\big\|_2$, where $N$ is a large number to push the indicator towards $0$ or $1$. We experimentally observed that $N = 10$ can ensure $\boldsymbol \alpha \in \{0,1\}^c$. (2) $\mathcal{L}_2 = ReLU(\|\boldsymbol \alpha_t\|_1/c-\tau_t)$, where $ReLU(\cdot)$ denotes the rectified linear activation function. It is easy to know that $\mathcal{L}_2 = 0$ if $\|\boldsymbol \alpha_t\|_1/c < \tau_t$, and $\mathcal{L}_2 = \|\boldsymbol \alpha_t\|_1/c - \tau_t$, otherwise. $\mathcal{L}_2$ is used to control the retention rate of preserved nodes. Accordingly, the compressed network is trained by:
\begin{align}
   \mathop{\arg\min}_{\theta, \theta^{'}, \boldsymbol s_t} &  \mathcal{L}_{c}(\boldsymbol{\hat{y}}^{(t)},\boldsymbol y; \theta, \theta^{'})+\lambda_1 \mathcal{L}_d(\boldsymbol{\hat{y}}^{(t)},\boldsymbol{\hat{y}}^{(t-1)}; \theta, \theta^{'})\\ \notag
   &+\lambda_2\big(\mathcal{L}_1(\boldsymbol s_t)+\mathcal{L}_2(\boldsymbol s_t)\big),
\end{align}
where $\lambda_2$ is a regularization factor.

The learnable compression approach endows the network with the adaptability to novel class samples; however, the results are largely impact by the initialization of $s_t$. Specifically, under the constraint of $\ell_1$ loss, when the initialization is smaller than zero, $s_t$ approaches $-N$ after network training; otherwise, $s_t$ approaches $+N$. This implies that whether to remove one node is simply up to the initialization of the learnable parameter.

\subsection{Node Self-activation}



To implement adaptive node compression, it is necessary to redefine the indicator $\boldsymbol \alpha_t$ rather than directly using the learnable parameter. Considering that network expansion is to enlarge the representation capacity, which is reflected by the features, one reasonable way for network compression is to make the indicator dependent on the outputs of nodes. 

To this end, we determine the indicator $\boldsymbol \alpha_t$ by introducing a self-activation mechanism, where the indicator vector $\boldsymbol \alpha_t$ is calculated as:
\begin{align}
   \boldsymbol \alpha_t = \frac{1}{1+e^{-\beta f(\boldsymbol x; \theta_m, \theta^{’(t)})}},
   \label{eq:alpha}
\end{align}
where $\beta = 1+epoch$ is an adjust rate to control the magnitude of the output $\boldsymbol \alpha_t$. As illustrated in Fig.\,\ref{fig:beta}, in the early learning epochs, $\beta$ is small and $\boldsymbol \alpha$ is a soft indicator between $0$ - $1$. The network gradient easily drives $\boldsymbol \alpha$ evolving from 0 to 1, or 1 to 0. When training proceeds, $\beta$ becomes large enough to force the indicator $\boldsymbol \alpha$ towards $0$ or $1$, eventually. Besides, whether to remove the added nodes depends on the outputs; that is, nodes with negative outputs are removed, and preserved otherwise. Moreover, the removed nodes might be adaptively recovered since the output $\mathbf{v}_{n_t}$ can dynamically change according to the network inputs.

If new classes are close to the old ones, few extra nodes are required to achieve good performance. Otherwise, more nodes are required. 
For an adaptive incremental learning, we propose to regard $\tau_t$ as a learnable parameter and define the following loss function to optimize it, as
\begin{align}
   \mathcal{L}_R= ReLU(\frac{\|\boldsymbol \alpha_t\|_1}{c}-\tau_t).
   \label{eq:L_R}
\end{align}
Combining Eq.\ \ref{eq:L_R} with Eq.\ \ref{eq:loss} results in the overall optimization objective for the LEC-Net, as
\begin{equation}
    \begin{split}
        \mathop{\arg\min}_{\theta, \theta^{'}, \tau_t} & \mathcal{L}_{c}(\boldsymbol{\hat{y}}^{(t)},\boldsymbol y; \theta, \theta^{'})+\lambda_1 \mathcal{L}_d(\boldsymbol{\hat{y}}^{(t)},\boldsymbol{\hat{y}}^{(t-1)};\theta, \theta^{'}) \\ & + \lambda_2 \mathcal{L}_R(\boldsymbol \alpha_t; \theta^{'}, \tau_t).
    \end{split}
  \label{eq:sa_loss}
\end{equation}

Optimizing Eq.\ \ref{eq:sa_loss} pursues a proper value for $\tau_t$ such that the expanded network reaches a trade-off between training performance (with respect to $\mathcal{L}$) and representation compactness (with respect to $\mathcal{L}_R$). When $\frac{\|\boldsymbol \alpha_t\|_1}{c} > \tau_t$, the gradient drives $\boldsymbol \alpha_t$ towards a sparse vector, which thus removes redundant nodes. When $\frac{\|\boldsymbol \alpha_t\|_1}{c} \leq \tau_t$, $\nabla {L}_R=0$, which means that $\tau_t$ stops updating and the number of expanded node becomes stable, while the ongoing optimization of $\mathcal{L}$ continuously boosts the network performance.

\begin{figure}[t]
    \centering
    \includegraphics[width=1\linewidth]{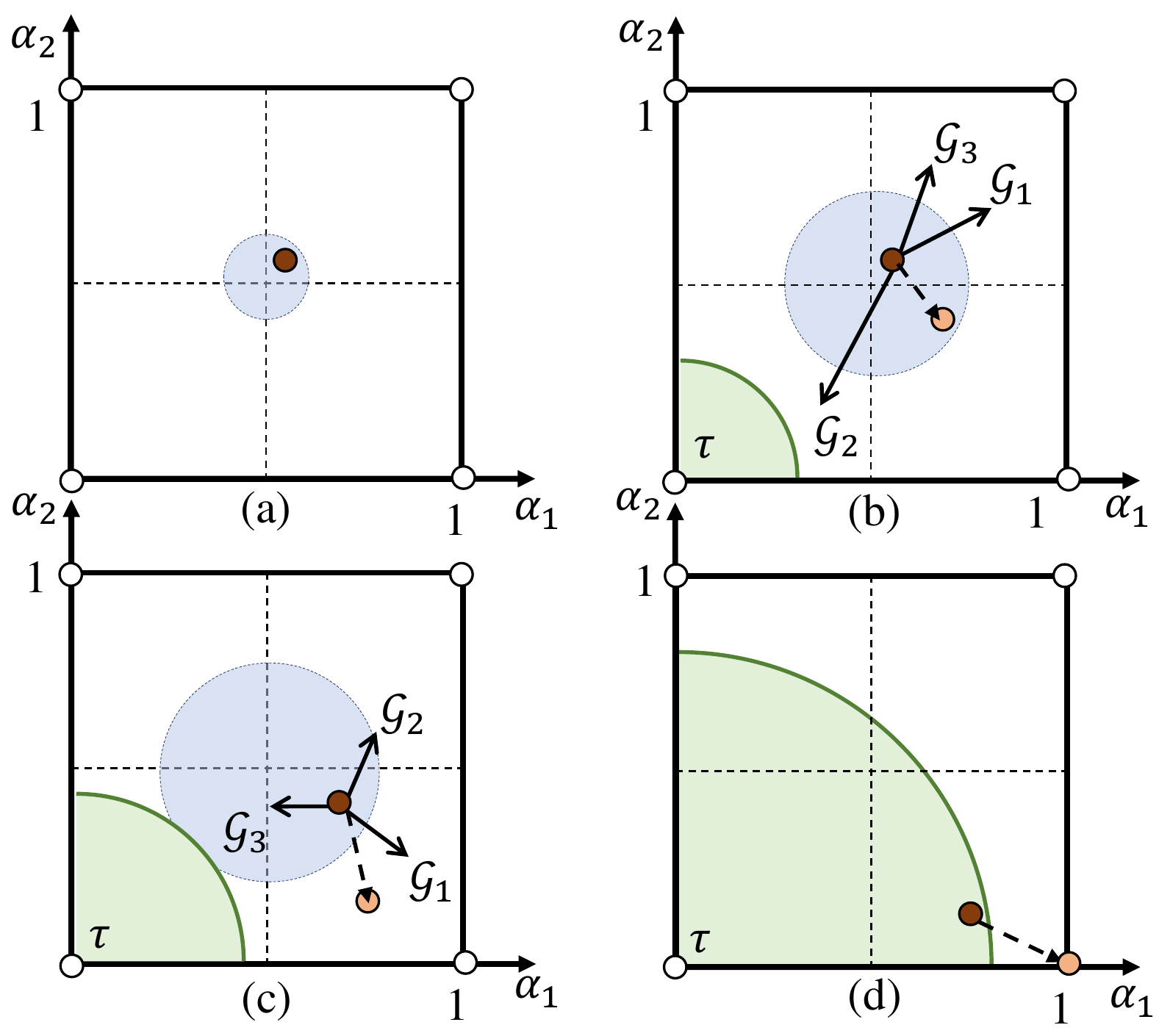}
    \caption{Illustration of the tentative optimization mechanism.}
    \label{fig:fig3}
\end{figure}

\subsection{Tentative Optimization}
The network expansion and compression procedure is formulated as a tentative optimization mechanism, where the objective (defined by Eq.\,\ref{eq:sa_loss}) is searching for the optimal indicator $\boldsymbol \alpha \in \{0, 1\}^c$.

To this end, the tentative network expansion firstly constructs a large optimization space which is then initialized by the node outputs. 
Denote the optimization space as $\mathcal{A}$, and $\boldsymbol \alpha=(\alpha_1, ..., \alpha_i, ..., \alpha_c) \in \mathcal{A}$ with $\alpha_i \in [0,1]$.
As shown in Fig.~\ref{fig:fig3}, each vertex $\boldsymbol \alpha \in \{0,1\}^c$ denotes a  solution. $\boldsymbol \alpha \in \{0\}^c$ and $\boldsymbol \alpha \in \{1\}^c$ respectively denote the network without expansion and 
network with expanded nodes. $\boldsymbol \alpha$ is initilized as $\boldsymbol \alpha = (1/c+\varepsilon_1,..., 1/c+\varepsilon_c)$.
The learnable compression procedure searches for the optimal $\boldsymbol{\alpha} \in \mathcal{A}$. 
Denote $\mathcal{G}_1$ as the variance of $\boldsymbol{\alpha}$ caused by $\beta$ and $\mathcal{G}_2, \mathcal{G}_3$ as the gradients of $\mathcal{L}_c, \mathcal{L}_R$ with respect to $\boldsymbol{\alpha}$, respectively. 
As shown in Fig.~\ref{fig:fig3}, $\mathcal{G}_1$ pushes $\boldsymbol \alpha$ to the vertex of the optimization space; $\mathcal{G}_2$ pursues an optimal value for $\boldsymbol \alpha$ so that the network learns new classes without forgetting old classes; $\mathcal{G}_3$ prevents $\boldsymbol \alpha$ from overfitting when $\boldsymbol \alpha$ falls out of the region of $\tau$.
In the early training epochs, $\boldsymbol \alpha$ is optimized in a small region by $\mathcal{G}_2, \mathcal{G}_3$ to search for a better direction. 
As training proceeds, $\boldsymbol \alpha$ is pushed by $\mathcal{G}_1$ to be away from the initial point. As a result, $\boldsymbol \alpha$ falls into the region of $\tau$ so that  the effect of $\mathcal{G}_3$ disappears, $\mathcal{G}_2$ is smaller than $\mathcal{G}_1$, and $\boldsymbol \alpha$ reaches one of the vertexes, which the sparsity of $\boldsymbol \alpha$ for network compression.

\section{Experiments}

\subsection{Experimental Setting}
\textbf{Datasets.}
We evaluate LEC-Net on three commonly used datasets including CIFAR100~\cite{CIFAR100}, CUB200~\cite{CUB200} and miniImageNet~\cite{MatchingNet}. For few-shot learning, categories in the datasets are divided into base ones with adequate annotations and novel ones with $K$-shot annotated images. For few-shot class-incremental learning (FSCIL), the network is trained upon the base classes for the first session. The novel classes are divided into $T$ learning sessions with $N$-way classes for each session for incremental learning, \emph{i.e.}, one base learning session and $T$ novel learning session. CIFAR100 and miniImageNet consist of 100 classes totally. We choose 60 of them as base classes and 40 as novel classes. Each novel class has 5 annotated images ($K=5$). The novel classes are divided into 8 sessions ($T=8$), each of which has 5 classes ($N=5$). CUB200 contains 200 classes where half are set as base classes and the other half as novel classes under the settings of $K=5$, $T=10$, $N=10$.

\textbf{Implementation Details.}
The baseline is built upon a simple network optimized by Eq.~\ref{eq:loss}, adopting Resnet18 as the backbone for a fair comparison with the state-of-the-art approach (TOPIC~\cite{TOPIC}). The code is implemented with PyTorch 1.0 and run on a Nvidia Tesla V100 GPU. During training, four data augmentation strategies, including normalization, horizontal flipping, random cropping, and random resizing, are used. The network is optimized with the SGD algorithm.

For the first learning session, we train the network using the dataset $D^{(0)}$ upon the base classes, with a batch size of 128 and an initial learning rate of 0.1. The learning rate is decreased to 0.01 after 60 epochs and stopped at the 100-$th$ epoch. When $t>0$, the network is trained by dataset $D^{(t)}$ with novel classes and the learning rate is set to 0.01. All the training images ($N \times K$) are fed to the network through a batch. The network stops training when the performance of the novel class reaches that of the old classes. Since the performance of the experiments is relevant to the class order and the labeled images of the novel classes, we conduct 10 times experiments by using different random seeds and report their average results.

\textbf{Evaluation.}
During inference, the network which is trained on the session $t$ with the dataset $D^{(t)}$ is evaluated on all of the seen class$\{C^{(0)}, C^{(1)},...,C^{(t)}\}$ under the metric of $ACC=\frac{TP+TN}{TP+TN+FP+FN}$ where TP, TN, FP and FN respectively denote the number of true positive, true negative, false positive and false negative predictions. 
Without specified, the experiment results refer to performance of the last session where all incremental classes are used.

\setlength{\tabcolsep}{4pt}
\begin{table}[t]
\begin{center}
\caption{Ablation study of LEC-Net on the CUB Dataset using the Resnet18 backbone. ``NE'' denotes the tentative note expansion module. ``NC'' denotes the learnable network compression module. ``SA'' denotes the self-activation module. }
\label{table:Ablation_Module}
\begin{tabular}{ccccc}
\hline\noalign{\smallskip}
Baseline & LEC (NE) & LEC (NC) & LEC (SA) & ACC\\
\noalign{\smallskip}
\hline
\checkmark & & & & 24.31\\
{}& \checkmark& & & 28.41\\
{}& & \checkmark & & 29.48\\
{}& & & \checkmark & \bf 31.96\\
\hline
\end{tabular}
\end{center}
\end{table}
\setlength{\tabcolsep}{1.4pt}

\setlength{\tabcolsep}{4pt}
\begin{table}[t]
\begin{center}
\caption{Ablation study by expanding and compressing different network layers on CUB Dataset with the Resnet18 backbone. ``FC'' denotes the fully connected layer.}
\label{table:Ablation_Module_2}
\begin{tabular}{cccccccclclclclclclcl}
\hline\noalign{\smallskip}
Net Layer& $3\times3$ conv. & $1\times1$ conv. & FC\\
\noalign{\smallskip}
\hline
ACC& 31.05 & 29.76 & \bf 31.96 \\
\hline
\end{tabular}
\end{center}
\end{table}
\setlength{\tabcolsep}{1.4pt}

\setlength{\tabcolsep}{4pt}
\begin{table}[t]
\begin{center}
\caption{Ablation study of the coefficient $\gamma$ on CUB Dataset with the Resnet18 backbone.}
\label{table:Ablation_gamma}
\begin{tabular}{cccccccccc|c}
\hline\noalign{\smallskip}
$\gamma$ & 0.2 & 0.4 & 0.6 & 0.8 & 1.0 & 1.2 & 1.4 & 1.6\\
\noalign{\smallskip}
\hline
ACC &23.8 &28.9 & 30.1& \bf 30.2&29.2&26.7 &24.3 &22.4\\
\hline
\end{tabular}
\end{center}
\end{table}
\setlength{\tabcolsep}{1.4pt}

\begin{figure}[t]
    \centering
    \includegraphics[width=1\linewidth]{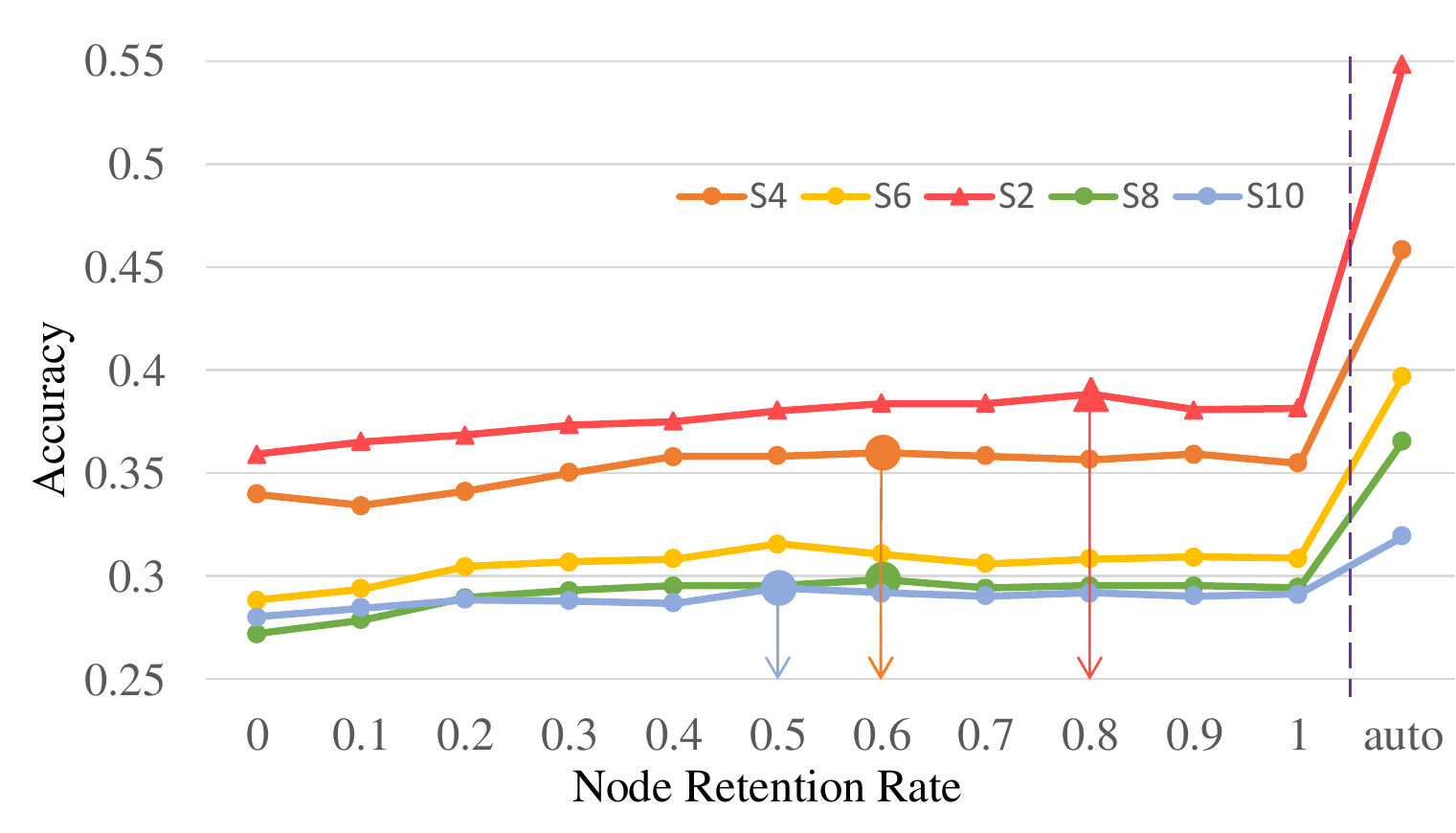}
    \caption{Performance of sessions under node retention rates.}
    \label{fig:ab_percentage}
\end{figure}

\begin{figure}[t]
    \centering
    \includegraphics[width=1\linewidth]{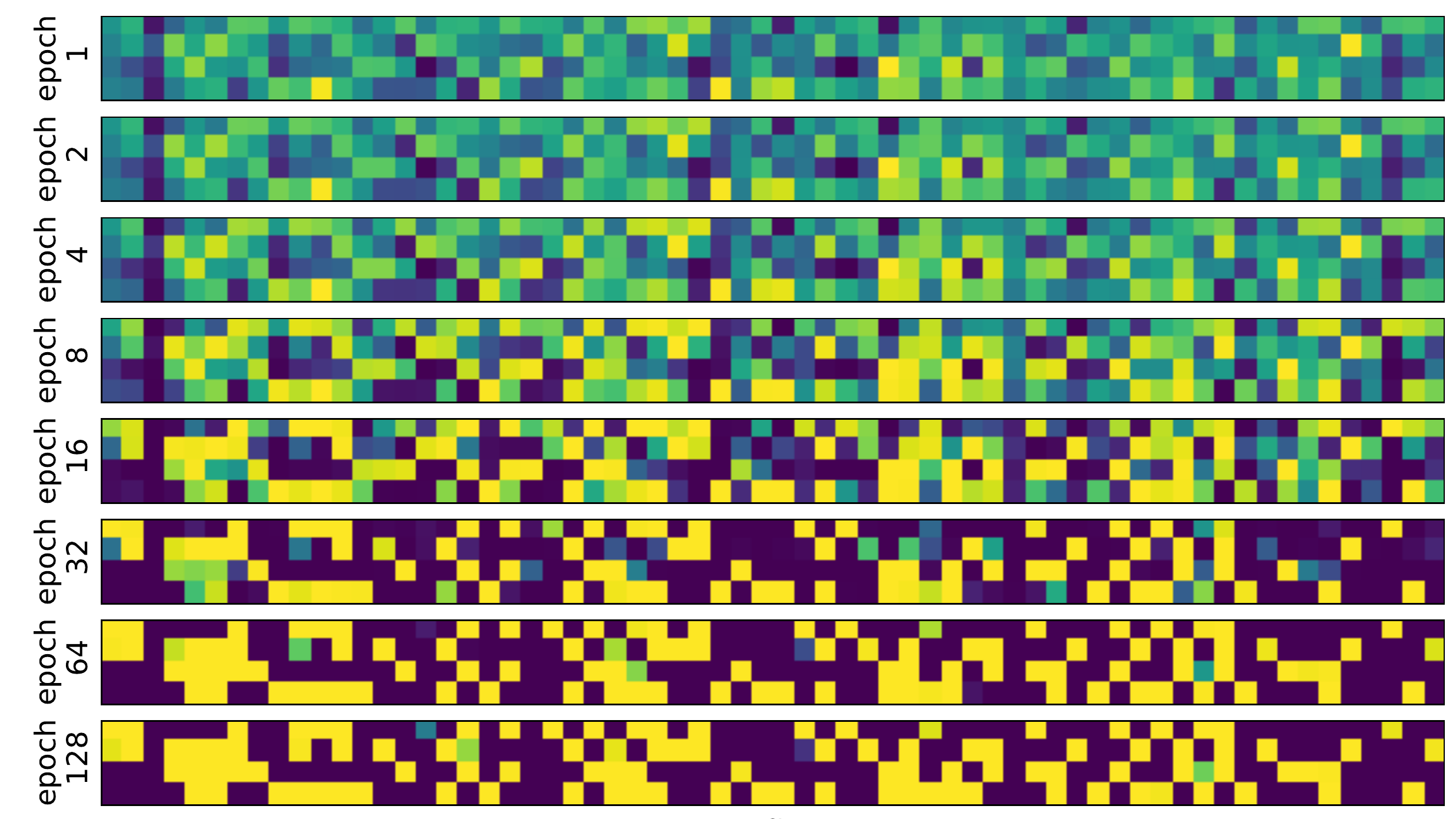}
    \caption{Visualization of the indicator $\boldsymbol \alpha$ during training. $\boldsymbol \alpha$ is reshaped to a $w\times h$ map for visualization.}
    \label{fig:alpha}
\end{figure}

\begin{figure}[t]
    \centering
    \includegraphics[width=1\linewidth]{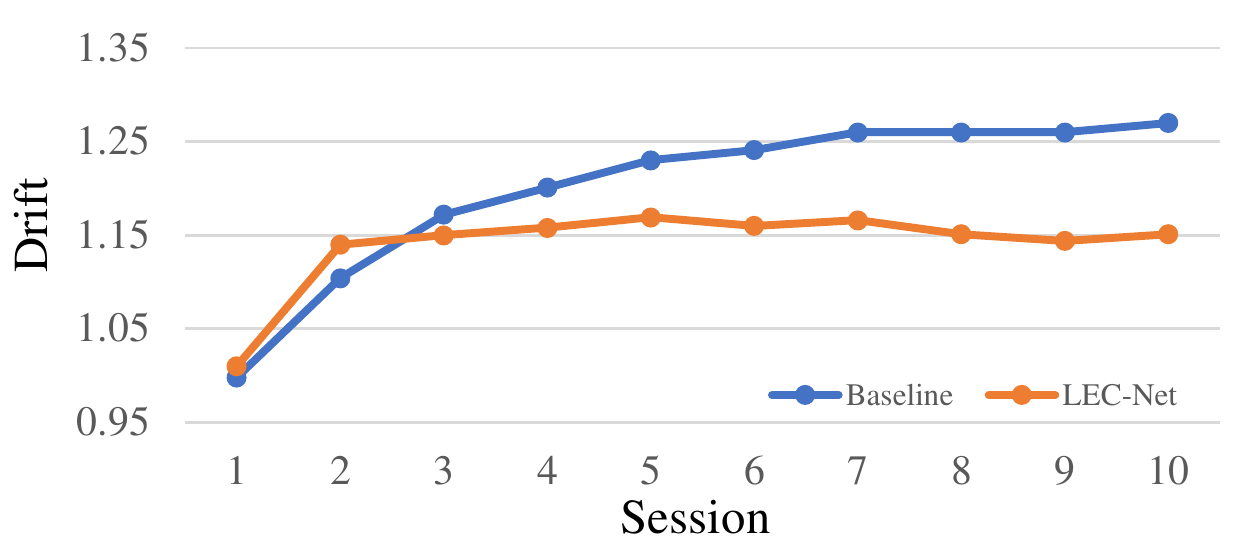}
    \caption{Comparison of feature drift by the baseline method and the proposed LEC-Net.}
    \label{fig:drift}
\end{figure}

\subsection{Ablation Study}
Table\ \ref{table:Ablation_Module} shows the efficacy of LEC-Net components at the last session for the CUB dataset. With network expansion, the average performance gain is $4.10\%$ (28.41\% \emph{v.s.} 24.31\%), indicating the significant advantage of network parameter increase for class incremental learning.
By Network compression, the performance gain increases to 5.17\% (29.48\% \emph{v.s.} 24.31\%). This shows that the network expansion introduces redundant parameters, which cause overfitting to novel classes. The introduction of network compression significantly alleviates such overfitting.
With the self-activation module for network compression, the performance gain increases to 7.65\% (31.96\% \emph{v.s.} 24.31\%). Since self-activation can dynamically change the indicator according to the network input, it improves the performance on both base and novel classes. 

\textbf{Coefficient $\gamma$.} In Table \ref{table:Ablation_gamma}, ablation study is carried out to determine the coefficient $\gamma$ to balance the old network feature and the expanded network feature. The best performance occurs at the extent of  $0.6\leq\gamma\leq 1.0$ and it shows that $\gamma=0.8$ reports the best performance. 
Very small values for $\gamma$ could exclude the feature representation learning by the old network while reducing the representation capability of overall features.
Very large values of $\gamma$ could consolidate the effect of the old network substantially increasing the risk of over-fitting to novel classes.

\begin{figure}[t]
    \centering
    \includegraphics[width=1\linewidth]{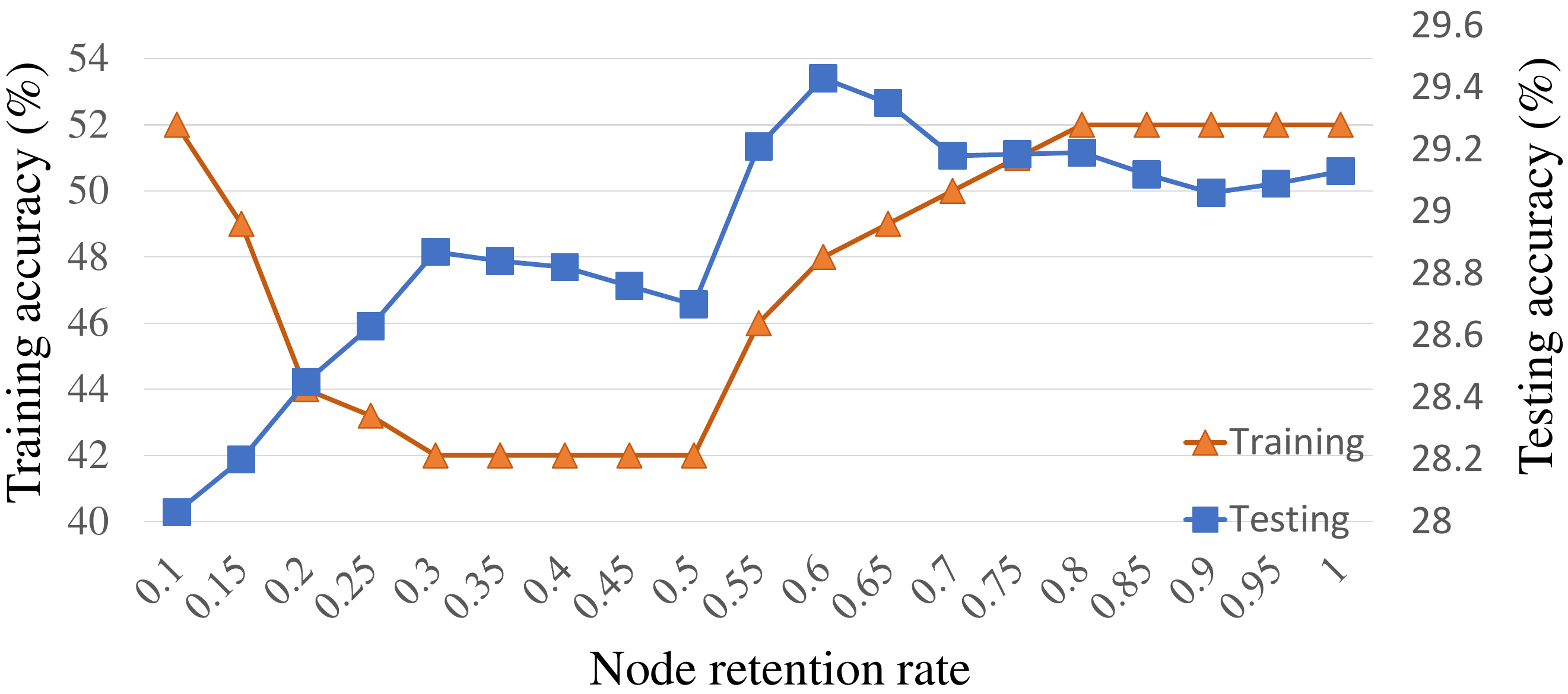}
    \caption{Training and testing performance under node retention rates.}
    \label{fig:retention}
\end{figure}

\textbf{Node Retention Rate ($\tau$)}. Experiments are conducted to explore the ablation of the node retention rate of the network.  Fig.~\ref{fig:ab_percentage} shows that $\tau$ performs differently on the different sessions. For example, for the second session, $\tau=0.8$ reports the best results. For the fourth session, $\tau=0.6$ performs the best. 
This shows that different novel class distributions requires different network expansion. 
If the novel classes are similar with the old classes, it requires fewer extra nodes to expand the representation capacity.
Otherwise, it requires more extra nodes.
By setting $\tau$ as a learnable network parameter, the network can specify additional network nodes for novel classes.

\textbf{Network layer}. Table~\ref{table:Ablation_Module_2} evaluates the performance by applying LEC-Net to different network layers, $e.g.,$ the $3\times3$ conv. layer, $1\times1$ conv. layer and the fully connected layer. Experimental results shows that LEC-Net achieves stable performance for different kinds of network layers. The best results are from the fully connected (fc) layer, because it is closest to the feature representation and therefore can be better optimized. 

\subsection{Model Analysis}

Fig.~\ref{fig:alpha} visualizes the evolution of the indicator $\alpha$ during incremental learning stages ($t>1$). In early epochs, $\boldsymbol \alpha_t$ is a soft indicator between 0-1 and the gradient drives it changing from 0 to 1 or from 1 to 0. When training proceeds and $\beta$ becomes larger, $\boldsymbol \alpha$ tends to $\{0,1\}^c$ and $\|\boldsymbol \alpha\|_1$ saturated to $\tau$.
Finally, $\alpha$ become a sparse vector indicating the preservation or removal of nodes.

Fig.~\ref{fig:drift} compares the feature drift of the baseline method and our LEC-Net approach. To quantify the feature drift, we define $Drift=\arccos{(\boldsymbol f^{(0)\mathsf{T}} \boldsymbol f^{(t)})}$, where $\boldsymbol f^{(0)\mathsf{T}}$ and $\boldsymbol f^{(t)}$ are the features trained by the $0$-$th$ session and $t$-$th$ session respectively. It can be seen that when incremental learning proceeds, feature drift of the baseline method increases significantly while that of LEC-Net becomes stable.

Fig.~\ref{fig:retention} shows the training and testing accuracy with different node retention rates. With the increase of note retention rate, the training accuracy firstly goes down and then goes up, while the testing accuracy firstly goes up and then goes down. The result indicates that  appropriate retained node rates between 0 and 1 can alleviate the overfitting problem. LEC-Net can optimize the retained node rate to achieve the best performance.

\begin{figure}[t]
    \centering
    \includegraphics[width=1\linewidth]{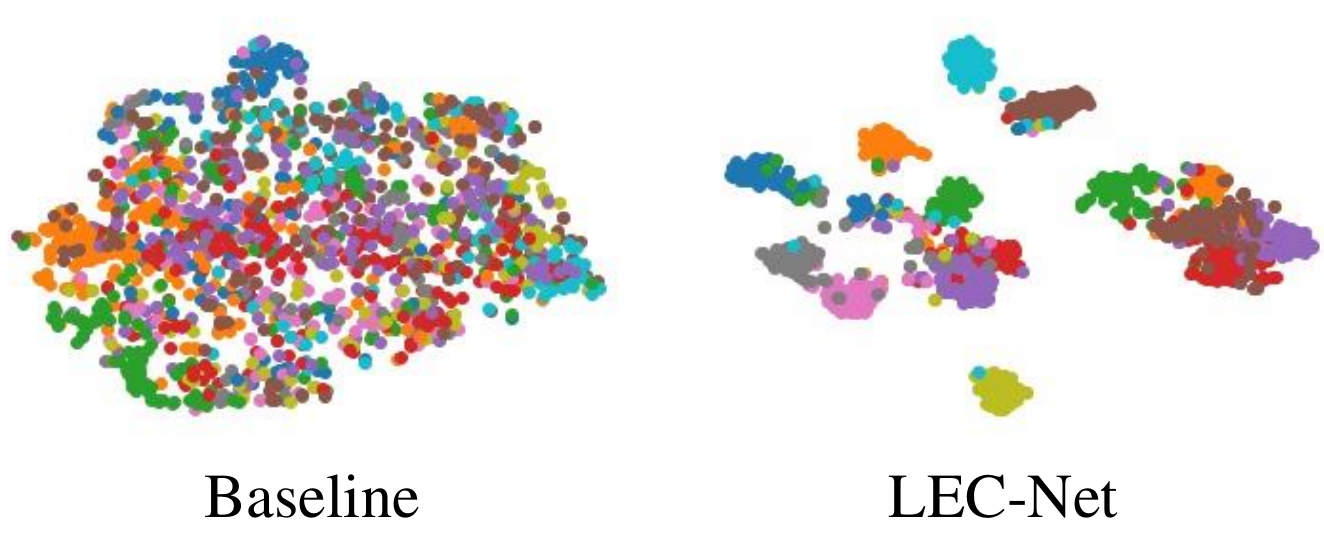}
    \caption{t-SNE visualization of the features in the last session.}
    \label{fig:Tsne}
\end{figure}

Fig.~\ref{fig:Tsne} visualizes the feature distributions of the baseline method and the proposed LEC-Net by respectively sampling 1024 images from base classes and new classes. One can see that the baseline method mixes up the base and novels classes while LEC-Net clearly separates these classes. This shows that the LEC-Net facilities optimizing the feature representation, which and reduce the overfitting problem during incremental learning.

\begin{table*}[t]
\begin{center}
\caption{Performance comparison on CUB with the Resnet18 backbone.}
\renewcommand\tabcolsep{6.0pt}
\label{table:SOTA_CUB}
\begin{tabular}{c|ccccccccccc}
\hline
\hline\noalign{\smallskip}
\multirow{2}{*}{Method} & \multicolumn{11}{c}{sessions} \\
& 0 & 1 & 2 & 3 & 4 & 5 & 6 & 7 & 8 & 9 &10\\
\hline
Ft-CNN &68.68 &44.81 &32.26&25.83 &25.62 &25.22 &20.84 &16.77 &18.82 &18.25&17.18\\
Joint-CNN &68.68 &62.43 &57.23&52.80 &49.50 &46.10 &42.80 &40.10 &38.70 &37.10&35.60\\
\hline
iCaRL~\cite{iCaRL} &68.68 &52.65 &48.61 &44.16 &36.62 &29.52 &27.83 &26.26 &24.01 &23.89 &21.16\\
EEIL~\cite{EEIL} &68.68 &53.63 &47.91 &44.20 &36.30 &27.46 &25.93 &24.70 &23.95 &24.13 &22.11\\
NCM~\cite{NCM} &68.68 &57.12 &44.21 &28.78 &26.71 &25.66 &24.62 &21.52 &20.12 &20.06 &19.87\\
TOPIC~\cite{TOPIC}  &68.68 &62.49 &54.81 &49.99 &45.25 &41.40 &38.35 &35.36 &32.22 &28.31 &26.28\\
\hline
\bf LEC-Net (ours)&70.86 &58.15 &\bf54.83 &49.34 &\bf45.85 &40.55 &\bf39.70 &34.59 &\bf36.58 &\bf33.56 &\bf31.96\\
\hline
\end{tabular}
\end{center}
\end{table*}
\setlength{\tabcolsep}{1.4pt}

\begin{table*}[t]
\begin{center}
\caption{Performance comparison on CIFAR100 with the Resnet18 backbone.}
\renewcommand\tabcolsep{8.0pt}
\label{table:SOTA_CIFAR}
\begin{tabular}{c|ccccccccc}
\hline
\hline\noalign{\smallskip}
\multirow{2}{*}{Method} & \multicolumn{9}{c}{sessions} \\
& 0 & 1 & 2 & 3 & 4 & 5 & 6 & 7 & 8\\
\hline
Ft-CNN & 64.10& 36.91& 15.37& 9.80& 6.67& 3.80& 3.70& 3.14& 2.65 \\
Joint-CNN &64.10& 59.30& 54.90& 51.20& 48.10& 45.80& 42.80& 40.90& 38.90\\
\hline
iCaRL~\cite{iCaRL} &64.10& 53.28& 41.69& 34.13& 27.93& 25.06& 20.41& 15.48& 13.73\\
EEIL~\cite{EEIL} &64.10& 53.11& 43.71& 35.15& 28.96& 24.98& 21.01& 17.26& 15.85\\
NCM~\cite{NCM} &64.10& 53.05& 43.96& 36.97& 31.61& 26.73& 21.23& 16.78& 13.54\\
TOPIC~\cite{TOPIC}  &64.10& 55.88& 47.07& 45.16& 40.11& 36.38& 33.96& 31.55& 29.37\\
\hline
\bf LEC-Net (ours) & 64.10 & 53.23 & 44.19 & 41.87 & 38.54 & \bf39.54 & \bf37.34 & \bf34.73 & \bf 34.73\\
\hline
\end{tabular}
\end{center}
\end{table*}
\setlength{\tabcolsep}{1.4pt}

\setlength{\tabcolsep}{4pt}
\begin{table}[t]
\begin{center}
\renewcommand\tabcolsep{2.0pt}
\caption{Performance on general incremental learning. *indicates the re-implemented performance.}
\label{table:general_inc}
\begin{tabular}{cccccc}
\hline
\hline
Method & SI~\cite{SI} & LwF~\cite{LwF}& DER*~\cite{DER} & LEC-Net\\
\hline
Accuracy & 19.27 & 19.62 & 19.61 & \bf19.90\\
\hline
\hline
\end{tabular}
\end{center}
\end{table}

\subsection{Performance}
Experiments and comparisons on the CUB200, CIFAR100 and miniImageNet datasets show that the proposed LEC-Net improves the state-of-the-art with significant margins.

\textbf{CUB200.} Table~\ref{table:SOTA_CUB} shows the performance on CUB200 with Resnet-18 backbone. 
It shows that LEC-Net achieves the best performance. Particularly, LEC-Net outperforms the NCM method~\cite{NCM} by 12.09\% (31.96\% \emph{v.s.} 19.87) and the TOPIC method~\cite{TOPIC} by 5.68\% (31.96\% \emph{v.s.} 26.28\%), which are significant margins for this challenging task. 
In early sessions, LEC-Net is on par with the TOPIC method~\cite{TOPIC}.
When learning proceeds and more classes are introduced, LEC-Net outperforms TOPIC significantly. This demonstrates the superiority of LEC-Net in handling the cases with significant class increasing but limited samples, which implies higher risk of feature drift and catastrophic forgetting.

\textbf{CIFAR100.} We report the performance on CIFAR100 dataset. Table~\ref{table:SOTA_CIFAR} shows that LEC-Net outperforms the state-of-the-arts by a large margin. Specifically, LEC-Net outperforms NCM~\cite{NCM} by 21.19\% (34.73\% \emph{v.s.} 13.54\%) and TOPIC~\cite{TOPIC} by 5.36\% (34.74\% \emph{v.s.} 29.37\%). In Table~\ref{table:SOTA_CIFAR} the performance of LEC-Net drops during the early sessions and then goes up. This is because the expansion of network during the early sessions leads to over-fitting on the CIFAR100 dataset, which causes performance drop. With the learnable compression mechanism which dynamically adjusts the nodes of networks to fit the increased classes, the network achieves better performance during the last few sessions, validating the adaptability of LEC-Net.

\textbf{General Incremental Learning.}
As a plug-and-play module, LEC-Net can be fused with a deep learning framework for general class incremental learning. By using DER~\cite{DER} (without memory buffer) as a baseline, we implement general incremental classification. The experiments are conducted on the Seq-Mnist dataset. Table~\ref{table:general_inc} shows that LEC-Net achieves the best performance. Specifically, LEC-Net outperforms LwF by 0.28\% (19.9\% \emph{v.s.} 19.62\%) and DER by 0.29\% (19.9\% \emph{v.s.} 19.61\%). Considering the small room for performance improvement on this dataset, the performance gains validate LEC-Net's potential for general incremental learning.

\section{Conclusion}
We proposed a learnable expansion-and-compression network (LEC-Net), and alleviated catastrophic forgetting and model over-fitting problems in a unified framework. By tentatively expanding network nodes, LEC-Net enlarged the representation capacity of features, reducing feature drift of old networks from the perspective of model regularization. By compressing the expanded network nodes, LEC-Net implemented minimal increase of parameters, alleviating over-fitting of the expanded network from a perspective of compact representation. While LEC-Net significantly improved the performance of FSCIL, as well as demonstrating the potential to be a general incremental learning approach with dynamic model expansion capability.

{\small
\bibliographystyle{ieee_fullname}
\bibliography{main}
}

\clearpage
\appendix
\section{Appendix}
\subsection{Discussion on Network Compression}

We analyze why the loss function defined by Eq.~\textcolor{red}{5} can make the indicator dependent on the initialization of $\boldsymbol s$. 
Revisit that the indicator $\boldsymbol \alpha$ is calculated by a learnable parameter $\boldsymbol s$ as $\boldsymbol \alpha = \frac{1}{1+e^{- \boldsymbol s}}$, and suppose $\boldsymbol s = (s_1,...,s_i,...,s_c)$. The gradient of $s_i$ is optimized by three loss functions including $\mathcal{L}_1$, $\mathcal{L}_2$, and $\mathcal{L}_c$ in the paper. Note that $\mathcal{L}_d$ is the distillation loss supervised by the old classes output, thus it doesn't affect the indicator which is learnt by the new classes. The gradient of  $\mathcal{L}_1$ with respect to $s_i$ is calculated as:
\begin{equation}
\begin{split}
\frac{\partial \mathcal{L}_1}{\partial s_i} & = \frac{\partial \big\||\boldsymbol s|-N\big\|_2}{\partial s_i} = \frac{\partial (|s_i|-N)^2}{\partial s_i} \\& =2(|s_i|-N)sign(s_i),
\label{eq:L1}
\end{split}
\tag {\romannumeral1}
\end{equation}
where $sign(x)$ is the sign function, which returns +1 if the input $x > 0$, -1 if the input $x < 0$ and 0 otherwise.

For $\mathcal{L}_2$, if $\| \boldsymbol{\alpha} \|_1 /c \leq \tau$, $\frac{\partial \mathcal{L}_2}{\partial s_i}=0$; otherwise, the gradient is calculated as:
\begin{equation}
\begin{split}
\frac{\partial \mathcal{L}_2}{\partial s_i} & = \frac{\partial (\|\boldsymbol \alpha\|_1/c-\tau)}{\partial s_i} = \frac{\partial \alpha_i/c}{\partial s_i} \\& =\frac{1}{c} \sigma(s_i)(1-\sigma(s_i)),
\label{eq:L2}
\end{split}
\tag {\romannumeral2}
\end{equation}
where $\sigma(\cdot)$ is the sigmoid function.

Denote $\boldsymbol f = f(x;\theta)$ and $\boldsymbol f^{'} = f(x;\theta, \theta^{'})$ as the features generated by the old network and the expanded network respectively. Eq.~\ref{eq:compress} is rewritten as $\boldsymbol f^{''} = (\boldsymbol \alpha \otimes \boldsymbol f^{'}) \oplus \gamma \boldsymbol f$ and we have
\begin{equation}
\begin{split}
\frac{\partial \mathcal{L}_c}{\partial s_i} & = \frac{\partial \mathcal{L}_c}{\partial f^{''}} \frac{\partial f^{''}}{\partial \alpha} \frac{\partial \alpha}{\partial s_i} \\& = \frac{\partial \mathcal{L}_c}{\partial f^{''}} \cdot f^{'}_i \cdot \sigma(s_i)(1-\sigma(s_i)).
\label{eq:Lc}
\end{split}
\tag {\romannumeral3}
\end{equation}

On the one hand, the gradients of $\mathcal{L}_1$ and $\mathcal{L}_2$ (under the condition of $\| \boldsymbol{\alpha} \|_1 /c > \tau$) are related to the value of $s_i$ according to Eq.~\ref{eq:L1} and Eq.~\ref{eq:L2}. As can be seen, the two gradient types drive indicator towards either 1 if the initial value of $s_i > 0$; or 0 otherwise. 
Thereby, it fails to adaptively decide which nodes are more important. 
On the other hand, we experimentally observe that the impact of this gradient is trivial. 
As results, the indicator largely relies on the initialization of $\mathbf{s}$.




\subsection{Discussion on Tentative Optimization}
We denote $\alpha_i$ as an element of $\boldsymbol \alpha$ and we have $\alpha_i = \frac{1}{1+e^{-\beta f^{'}_i}} = \sigma(\beta f^{'}_i)$. 
Based on the definition of $\beta$, it is not relative to the node output $f^{'}_i$. The first order Taylor expansion of $\alpha_i$ from $\alpha_i^{(0)}$ is written as: $\alpha_i = \alpha_i^{(0)} + \frac{\partial \sigma(\beta f^{'}_i)}{\partial \beta}\triangle \beta + \frac{\partial \sigma(\beta f^{'}_i)}{\partial f^{'}_i}\triangle f^{'}_i$, where$\triangle f^{'}_i = \frac{\partial f^{'}_i}{\partial \theta^{'}}\triangle \theta^{'}$. $\theta^{'}$ is optimized by Eq.~\ref{eq:sa_loss} and $\triangle \theta^{'} = -(\frac{\partial \mathcal{L}_c}{\partial \theta^{'}} +\frac{\partial \mathcal{L}_R}{\partial \theta^{'}})$. We have
\begin{equation}
\begin{split}
    \triangle \alpha_i &= \frac{\partial \sigma(\beta f^{'}_i)}{\partial \beta}\triangle \beta+\frac{\partial \sigma(\beta f^{'}_i)}{\partial f^{'}_i}\frac{\partial f^{'}_i}{\partial \theta^{'}}(-(\frac{\partial \mathcal{L}_c}{\partial \theta^{'}} +\frac{\partial \mathcal{L}_R}{\partial \theta^{'}}))\\
    & = \mathcal{G}_1+\mathcal{G}_2+\mathcal{G}_3,
\end{split}
\tag {\romannumeral4}
\end{equation}
where $\mathcal{G}_1=\frac{\partial \sigma(\beta f^{'}_i)}{\partial \beta}\triangle \beta$ is the variance of $\alpha_i$ caused by $\beta$, $\mathcal{G}_2=\frac{\partial \sigma(\beta f^{'}_i)}{\partial f^{'}_i}\frac{\partial f^{'}_i}{\partial \theta^{'}}(-\frac{\partial \mathcal{L}_c}{\partial \theta^{'}})$, and $\mathcal{G}_3 = \frac{\partial \sigma(\beta f^{'}_i)}{\partial f^{'}_i}\frac{\partial f^{'}_i}{\partial \theta^{'}}(-(\frac{\partial \mathcal{L}_R}{\partial \theta^{'}})$ are the variance of $\alpha_i$ caused by the gradient of the $\mathcal{L}_c$ and $\mathcal{L}_R$ respectively.

\begin{table*}[!t]
\begin{center}
\caption{Performance Comparison on miniImageNet with Resnet18 backbone.}
\renewcommand\tabcolsep{8.0pt}
\label{table:SOTA_Miniimage}
\begin{tabular}{c|ccccccccc}
\hline
\hline\noalign{\smallskip}
\multirow{2}{*}{Method} & \multicolumn{9}{c}{sessions} \\
& 0 & 1 & 2 & 3 & 4 & 5 & 6 & 7 & 8\\
\hline
Ft-CNN & 61.31& 27.22& 16.37& 6.08& 2.54& 1.56& 1.93& 2.60& 1.40 \\
Joint-CNN &61.31& 56.60& 52.60& 49.00& 46.00& 43.30& 40.90& 38.70& 36.80\\
\hline
iCaRL &61.31& 46.32& 42.94& 37.63& 30.49& 24.00& 20.89& 18.80& 17.21\\
EEIL &61.31& 46.58& 44.00& 37.29& 33.14& 27.12& 24.10& 21.57& 19.58\\
NCM &61.31& 47.80& 39.31& 31.91& 25.68& 21.35& 18.67& 17.24& 14.17\\
TOPIC  &61.31& 50.09& 45.17& 41.16& 37.48& 35.52& 32.19& 29.46& 24.42\\
\hline
\bf LEC-Net (ours) & 61.31 & 35.37 & 36.66 & 38.59 & 33.90 & \bf 35.89 & \bf 36.12 & \bf 32.97 & \bf 30.55\\
\hline
\end{tabular}
\end{center}
\end{table*}
\setlength{\tabcolsep}{1.4pt}

More specifically, $\mathcal{G}_1$,$\mathcal{G}_2$,$\mathcal{G}_3$ is written as:
\begin{equation}
\begin{split}
    \mathcal{G}_1 & = \frac{\partial \sigma(\beta f^{'}_i)}{\partial \beta} \triangle \beta\\ & =\sigma(\beta f^{'}_i)(1-\sigma(\beta f^{'}_i))f^{'}_i\triangle \beta \\
    &=\zeta(\beta f^{'}_i) f^{'}_i \triangle \beta,
\end{split}
\tag {\romannumeral5}
\end{equation}
where $\zeta(\cdot)=\sigma(\cdot)(1-\sigma(\cdot))$. $\mathcal{G}_1$ pushes $\boldsymbol \alpha$ to the vertex of the optimization space which is related to the node output $f^{'}_i$. For example, when $\alpha_i>0.5$ ($f^{'}_i>0$), $\mathcal{G}_1>0$ which pushes $\alpha_i$ to 1. 
\begin{equation}
\begin{split}
    \mathcal{G}_2 =& -\frac{\partial \sigma(\beta f^{'}_i)}{\partial f^{'}} \frac{\partial f^{'}_i}{\partial \theta^{'}} \frac{\partial \mathcal{L}_c}{\partial \theta^{'}}\\
    =& -\beta \sigma(\beta f^{'}_i)(1-\sigma(\beta f^{'}_i)) \frac{\partial f^{'}_i}{\partial \theta^{'}}\\
    & \cdot \Big(\frac{\partial \mathcal{L}_c}{\partial f^{''}} \frac{\partial f^{'}}{\partial \theta^{'}} \big(\beta \sigma(\beta f^{'}_i)(1-\sigma(\beta f^{'}_i)) f^{'}_i+\sigma(\beta f^{'})\big) \Big)\\
    =& -\beta^2 (\zeta(\beta f^{'}_i))^2 (\frac{\partial f^{'}_i}{\partial \theta^{'}})^2 \frac{\partial \mathcal{L}_c}{\partial f^{''}} f^{'} - \beta \zeta(\beta f^{'}_i) \frac{\partial f^{'}_i}{\partial \theta^{'}} \sigma(\beta f^{'}_i).
\end{split}
\tag {\romannumeral6}
\end{equation}

As can be seen, the direction of $\mathcal{G}_2$ depends on the gradient of $\frac{\partial \mathcal{L}_c}{\partial f^{''}}$, the node output $f^{'}$ and the gradient of $\frac{\partial f^{'}_i}{\partial \theta^{'}}$. And $\mathcal{G}_2$ pursue an optimal value for $\boldsymbol \alpha$ so that the network learns new classes without forgetting old ones. If the direction of $\mathcal{G}_2$ for all of the node is negative, the optimal solution is $\{0\}^c$. We consider $\mathcal{G}_2>0$ for a detailed analysis.

If $\|\boldsymbol \alpha\|_1/c > \tau$:
\begin{equation}
\begin{split}
    \mathcal{G}_3 =& -\frac{\partial \sigma(\beta f^{'}_i)}{\partial f^{'}} \frac{\partial f^{'}_i}{\partial \theta^{'}} \frac{\partial \mathcal{L}_R}{\partial \theta^{'}}\\
    =& -\beta \sigma(\beta f^{'}_i)(1-\sigma(\beta f^{'}_i)) \frac{\partial f^{'}_i}{\partial \theta^{'}}\\ 
    & \cdot \frac{\lambda}{c}\frac{\partial f^{'}}{\partial \theta^{'}}\beta\sigma(\beta f^{'})(1-\sigma(\beta f^{'}))\\
    =& -\frac{\lambda}{c} \beta^2 (\zeta(\beta f^{'}_i))^2 (\frac{\partial f^{'}_i}{\partial \theta^{'}})^2.
\end{split}
\tag {\romannumeral7}
\end{equation}

If $\|\boldsymbol \alpha\|_1/c>\tau$, $\mathcal{G}_3<0$; otherwise, $\mathcal{G}_3=0$. It means that $\mathcal{G}_3$ prevents $\boldsymbol \alpha$ from overfitting when $\boldsymbol \alpha$ falls out of the region of $\tau$.  The parameters of tentative expansion network are initialized randomly, and $f^{'} = 0+\varepsilon$.

In the early training epochs, $|\mathcal{G}_1| << |\mathcal{G}_2|$ and $|\mathcal{G}_3|$ since $f^{'}_i \approx 0$ resulting in $\mathcal{G}_1 \approx 0$. $\boldsymbol \alpha$ is optimized in a small region by $\mathcal{G}_2$ and $\mathcal{G}_3$ to search for a better direction towards the vertex. 
As training proceeds, $\boldsymbol \alpha$ is pushed by $\mathcal{G}_1$ continually to be away from the initial point. During this period, $\beta$ grows up which leads to smaller $\zeta(\beta f^{'}_i)$ and $\beta \zeta(\beta f^{'}_i)$ \big($\zeta(\cdot)$ is a monotonously decreasing function if the input is larger than zero. Besides, $\frac{\partial \big(\beta \zeta(\beta f^{'}_i)\big)}{\partial \beta} < 0$ denoting that $\beta \zeta(\beta f^{'}_i)$ is also a monotonously decreasing function when $\beta$ becomes large\big). 
Besides, when training converges, $\frac{\partial \mathcal{L}_c}{\partial f^{''}}$ and $\frac{\partial f^{'}_i}{\partial \theta^{'}}$ decrease. As a result, the magnitude of $\mathcal{G}_1, \mathcal{G}_2, \mathcal{G}_3$ are decreasing together.

With $\boldsymbol \alpha$ getting away from the initial point continually, it falls into the region of $\tau$ so that the effect of $\mathcal{G}_3$ disappears. $\mathcal{G}_2$ is smaller than $\mathcal{G}_1$. This means that it is a stable solution of the optimization. As the result, $\boldsymbol \alpha$ reaches one of the vertexes, resulting in the network compression.

\subsection{Experiments on miniImageNet}
%

Table.~\ref{table:SOTA_Miniimage} shows the performance on miniImageNet with Resnet18 backbone. It shows that the proposed LEC-Net outperforms state-of-the-art TOPIC~\cite{TOPIC} by 6.13\% (30.55 \emph{v.s.} 24.42\%) and achieves the best performance. For session 1, the performance drops when the performance of the novel class reaches that of the old classes. Although TOPIC reports the best-performing one among all training epochs, its performance on novel classes are lower than that of old ones.

\subsection{General Incremental Learning}
Following the settings in DER~\cite{DER}, we divide the training
samples into give tasks, each of which contains 2 classes.
All the classes are fed to the network for model training in a fixed order across different runs.
%
For fair comparison, we remove the memory buffer mechanism by replacing the fully-connected layers of DER with LEC-Net to implement general incremental learning.

\end{document}